\documentclass[10pt,twocolumn,letterpaper]{article}

\usepackage{cvpr}              

\usepackage[dvipsnames]{xcolor}

\definecolor{cvprblue}{rgb}{0.21,0.49,0.74}
\usepackage[pagebackref,breaklinks,colorlinks,citecolor=cvprblue]{hyperref}
\newcommand{\delmetric}{\texttt{ReMOVE}\xspace}

\usepackage{pifont}
\usepackage{lipsum}
\newcommand{\cmark}{{\color{green}\ding{51}}}
\newcommand{\xmark}{{\color{red}\ding{55}}}
\usepackage{soul}
\usepackage{multirow}
\usepackage{textcomp}
\newcommand{\textapproximate}{\raisebox{0.5ex}{\texttildelow}}
\newsavebox{\measurebox}
\def\paperID{18} 
\def\confName{CVPR}
\def\confYear{2024}

\title{ReMOVE: A Reference-free Metric for Object Erasure}

\author{Aditya Chandrasekar$^{\,1, 2}$\textsuperscript{\textsection} \; Goirik Chakrabarty$^{\,1}$\textsuperscript{\textsection} \; Jai Bardhan$^{\,1}$ \; Ramya Hebbalaguppe$^{\,1, 3}$ \; Prathosh AP$^{\,2}$\\
$^{1}\,$TCS Research \quad $^{2}\,$IISc Bangalore \quad $^{3}\,$IIT Delhi\\
{\small \url{https://github.com/chandrasekaraditya/ReMOVE}}
}

\begin{document}

\maketitle

\begingroup\renewcommand\thefootnote{\textsection}
\footnotetext{Equal contribution.}
\endgroup

\begin{abstract}

We introduce \delmetric, a novel reference-free metric for assessing object erasure efficacy in diffusion-based image editing models post-generation. Unlike existing measures such as LPIPS and CLIPScore, \delmetric addresses the challenge of evaluating inpainting without a reference image, common in practical scenarios. It effectively distinguishes between object removal and replacement. This is a key issue in diffusion models due to stochastic nature of image generation. Traditional metrics fail to align with the intuitive definition of inpainting, which aims for (1) seamless object removal within masked regions (2) while preserving the background continuity. \delmetric not only correlates with state-of-the-art metrics and aligns with human perception but also captures the nuanced aspects of the inpainting process, providing a finer-grained evaluation of the generated outputs.

\end{abstract}

\section{Introduction}
\label{sec:intro}

In the contemporary creative landscape, diffusion models have surged in popularity, driving innovation in visual content generation~\cite{dall-e, stablediff, sora}. One of the primary applications of diffusion models lies in their role in image editing, achieved through prompt-based user inputs, which streamline access to complex editing functionalities in the creative process~\cite{lomoe, ip2p, bld, anydoor}. These models provide a robust methodology for various tasks, including object replacement, position switching, and object erasure~\cite{editval, piebench}. Among these categories, \textit{object erasure is the task of inpainting a masked region with neighbouring pixels to seamlessly remove undesired objects from an image.}

Image inpainting, a technique in computer vision akin to a digital paintbrush, serves to restore or complete images by intelligently reconstructing missing or damaged areas~\cite{lama, ffc}. This process involves filling in the gaps with details that seamlessly blend with the surrounding context. Moreover, inpainting extends to removing unwanted objects from a photo, such as a power line disrupting a scenic landscape. By analyzing the surroundings, the technique can reconstruct the missing pixels, creating an image as if the power line was never there.

\begin{figure}[t]
    \centering
    \includegraphics[width=\linewidth]{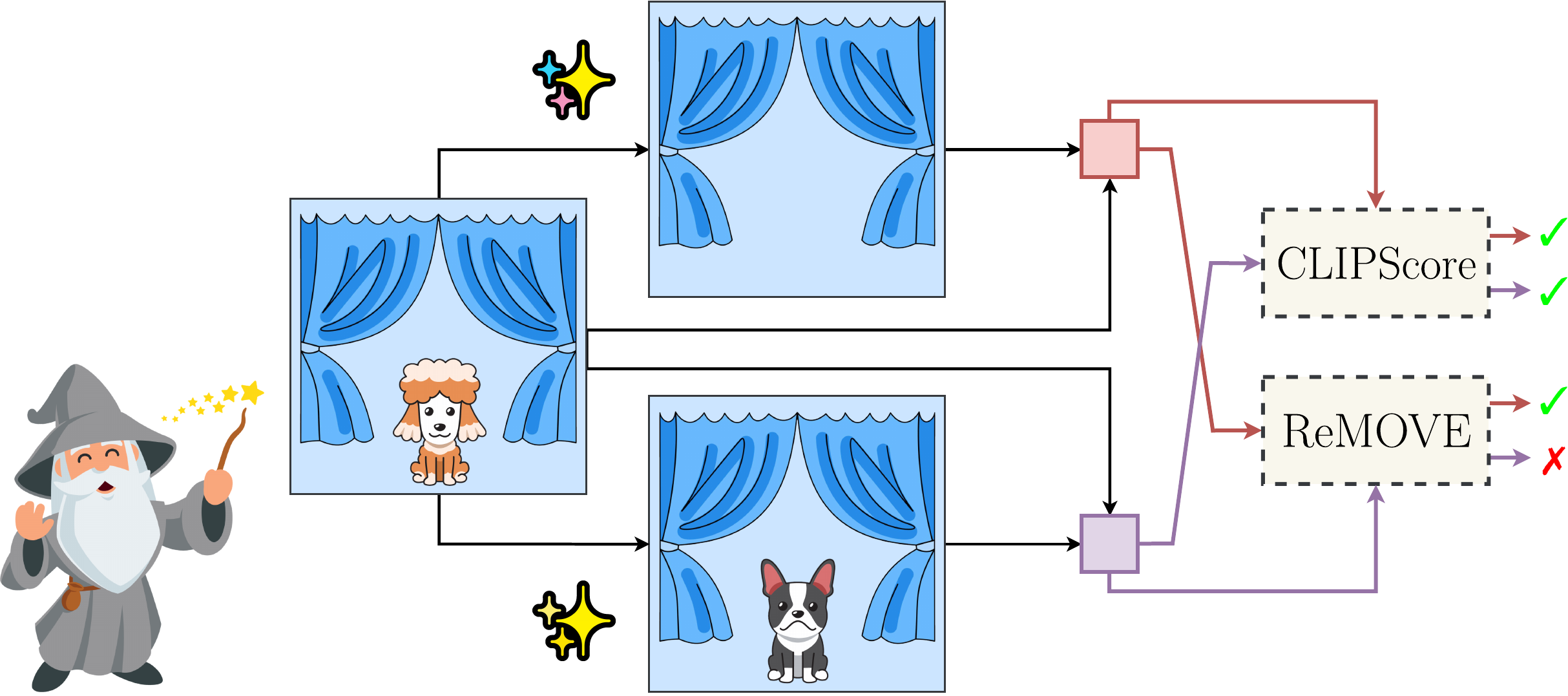}
    \caption{\textbf{Motivation for \delmetric}: Comparison of \delmetric with CLIPScore, illustrating the latter's lack of distinction (denoted by \cmark and \xmark) between two inpainting methods: method \includegraphics[height=0.009\textheight]{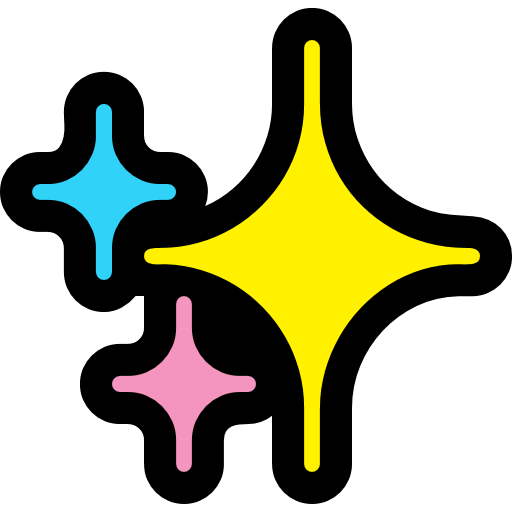}, which erases the object, and method \includegraphics[height=0.009\textheight]{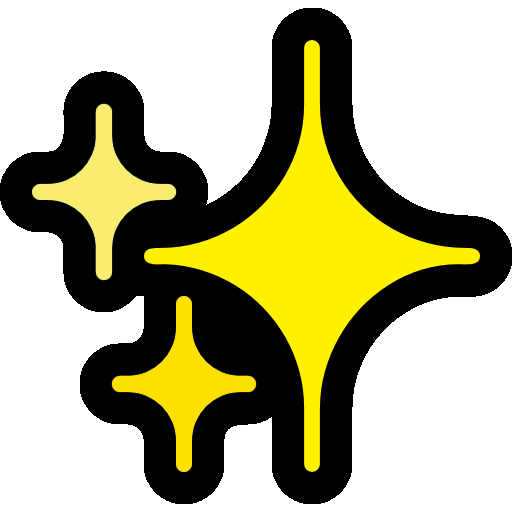}, which replaces it with another.}
    \label{fig:teaser}
\end{figure}

Over the years, there has been significant advancements in image inpainting~\cite{inpainting-survey} and it's evaluation metrics. These metrics typically fall into three categories: structure-based, saliency-based, and machine learning-based. However, the majority of these metrics are reference-dependent, requiring a ground truth inpainting, which is often very hard to obtain on a large scale for real images~\cite{metric-survey}. This reliance on reference-based evaluation can pose challenges, highlighting the need for developing alternative approaches to assess the quality and efficacy of inpainting methods. 

\begin{figure*}[t]
    \centering
    \includegraphics[width=\textwidth]{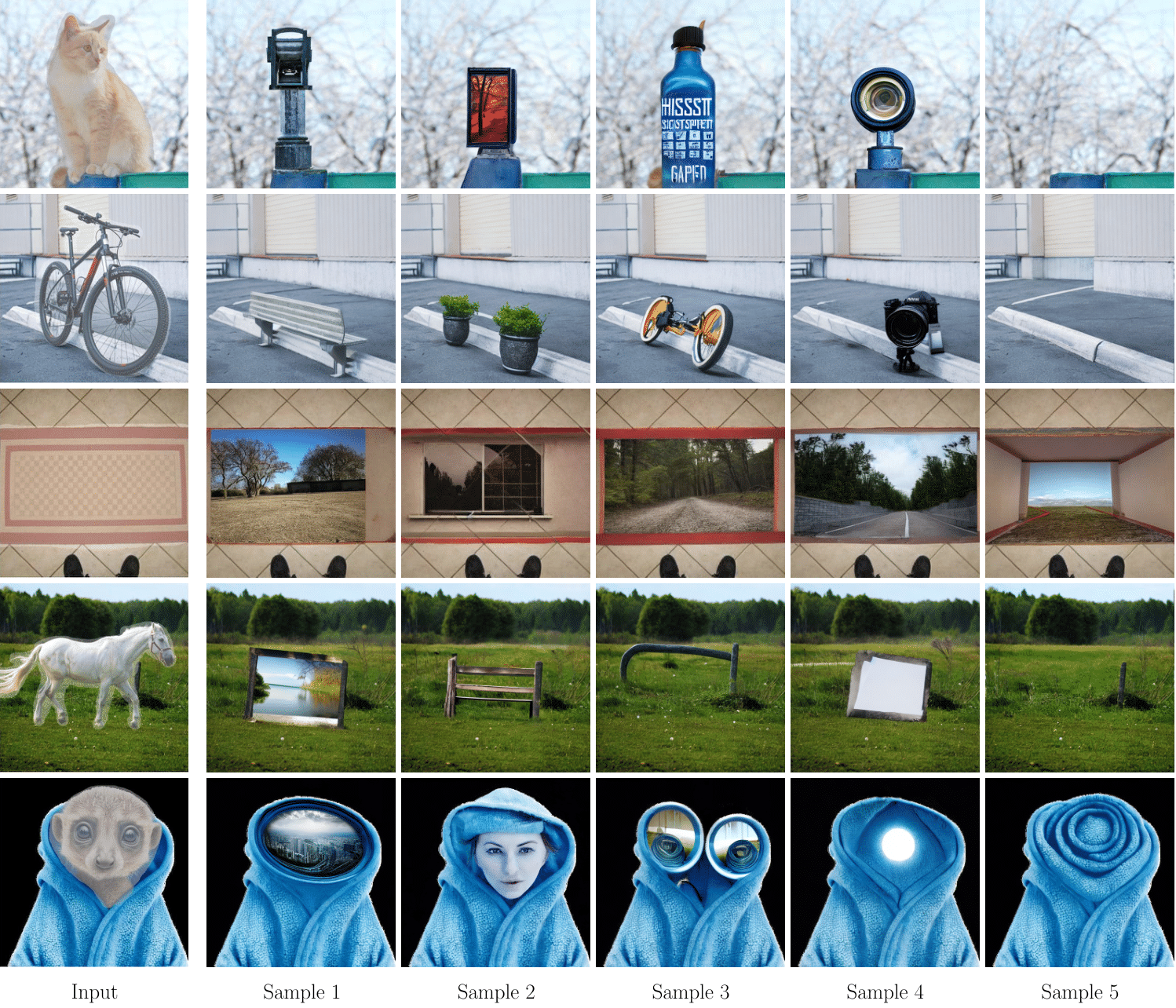}
    \caption{\textbf{Randomness in Object Inpainting using SD-Inpaint}: Samples of diffusion-based image inpainting using SD-Inpaint~\cite{stablediff} generated across varying seeds. The object intended for inpainting is substituted with a different object rather than replacing it with the background. In some cases (column 6), the model replaces the object with background pixels as desired.}
    \label{fig:sd-failure}
\end{figure*}

\begin{figure*}[t]
    \centering
    \includegraphics[width=\textwidth]{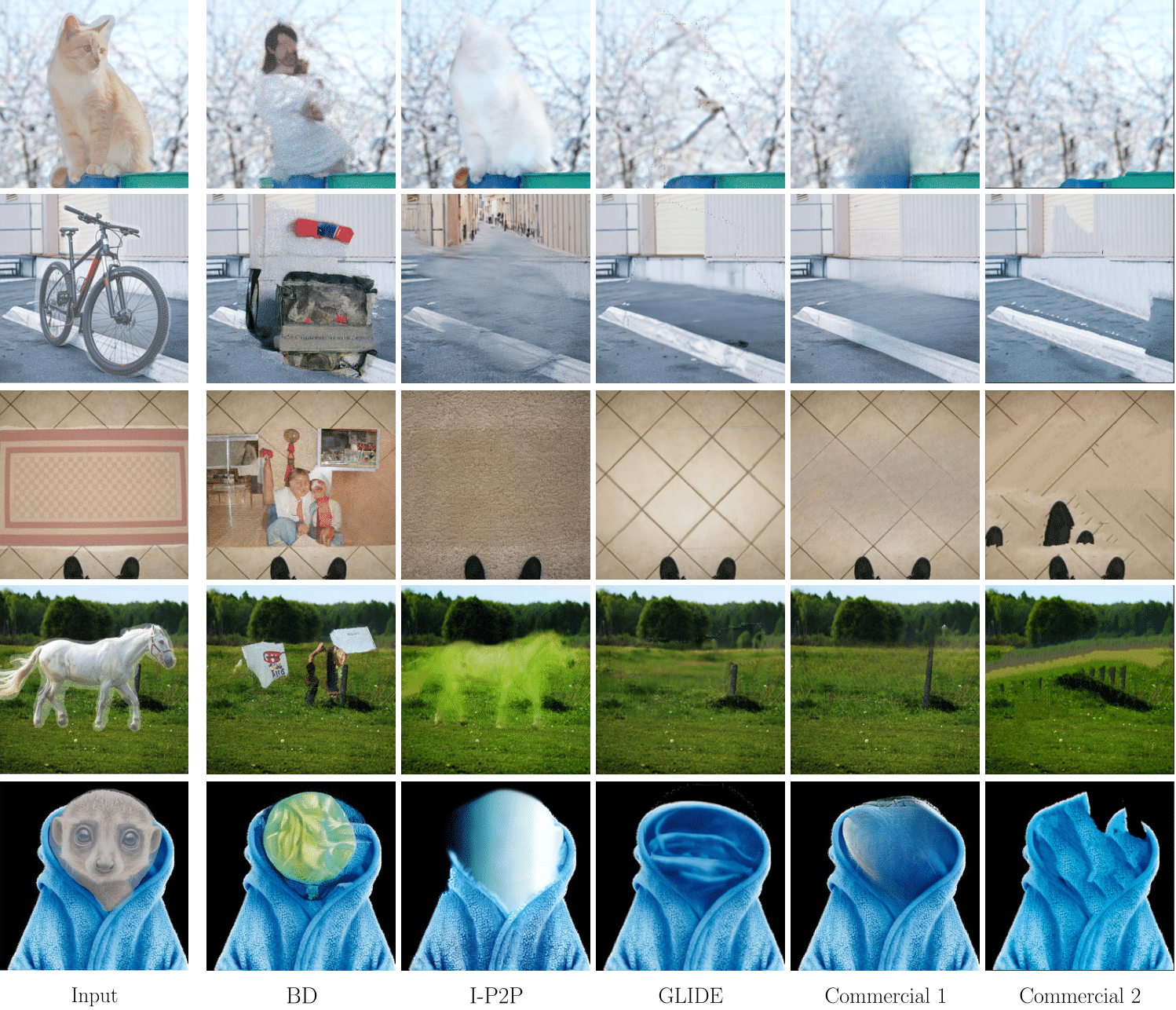}
    \caption{\textbf{Randomness in Object Inpainting with Other Methods}: Samples of diffusion-based image inpainting using multiple methods. The object intended for inpainting is often substituted with a different object rather than replacing it with the background.}
    \label{fig:other-methods}
\end{figure*}

\begin{figure*}[t]
    \centering
    \includegraphics[width=0.95\linewidth]{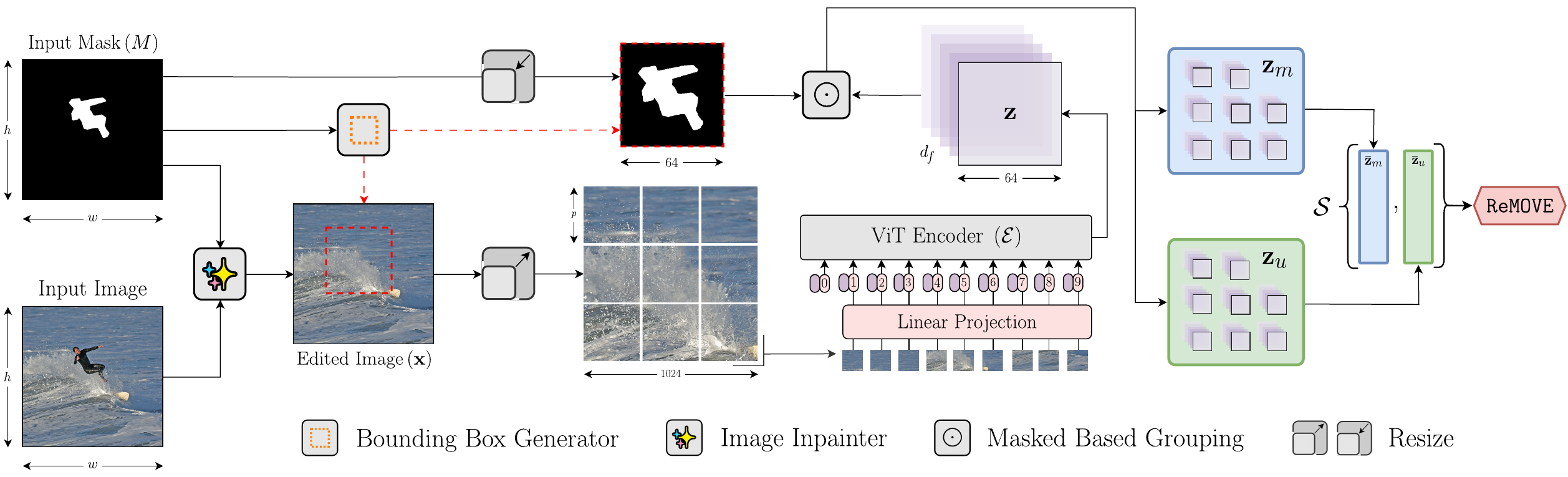}
    \caption{\textbf{Schematic Diagram of \texttt{ReMOVE}:} The inpainter \includegraphics[height=0.009\textheight]{img/Teaser/s1.png} takes the original image and (optionally) object mask to produce an edited image with the object deleted. Our metric requires only the edited image and object mask. After preprocessing using a bounding box crop (only in the crop-variant) and resizing, the image is tokenized into patches, and the encoder $\mathcal{E}$ obtains features for each patch. Simultaneously, the mask is resized and used to split patch embeddings into object and background embeddings. The mean feature embeddings are compared using a similarity measure to yield \delmetric.}
    \label{fig:schematic}
\end{figure*}

Zhang~\etal~\cite{lpips} show that traditional per-pixel metrics like Mean Squared Error (MSE) and Peak Signal-to-Noise Ratio (PSNR) often employed in image enhancement, super-resolution~\cite{phishnet,superres-metrics} are inadequate for evaluating structured outputs such as images. This is because these metrics assume independence among pixels, failing to capture perceptual changes accurately. A notable illustration is the discrepancy between perceptual differences and minor changes in MSE induced by blurring or noising~\cite{ssim}. Thus, the authors proposed Learned Perceptual Image Patch Similarity (LPIPS)~\cite{lpips}, obtained using feature distances in a neural network, outperforming traditional metrics while accounting for many nuances of human perception. Another such metric is the CLIPScore~\cite{clipscore}, which uses multi-modal deep features to quantify image-text similarity. LPIPS and CLIPScore have also been used to measure the quality of inpainting, but both of them need access to the ground truth inpainted image~\cite{diffusioneditingsurvey}. In the case of CLIPScore, the prompt can be obtained using a captioning model like BLIP~\cite{blip} on the ground truth. However, the reference-free version of CLIPScore~\cite{clipscore} only looks for the removal of the intended object and does not distinguish between object removal and object replacement, as illustrated in \cref{fig:teaser}.

In this work, we propose \delmetric, a reference-free metric to access inpainting quality, with a focus on object erasure using diffusion-based editing frameworks. This is primarily because these frameworks often introduce new objects alongside the removal of existing ones, as shown in \cref{fig:sd-failure}. The issue persists across other inpainting frameworks built on Stable Diffusion~\cite{bld,ip2p}, while other commercial inpainters~\cite{pincel, theinpaint} and inpainting models such as GLIDE~\cite{glide} demonstrate superior performance in addressing this challenge, as seen in~\cref{fig:other-methods}. To assess quality, we measure the distance between mean patchwise features of the masked and unmasked regions obtained from a Vision Transformer (ViT)~\cite{vit}, pre-trained on a segmentation task~\cite{sam}. We are inspired by Lugmayr \etal \cite{repaint}, wherein they train a model to inpaint a masked region conditioned on the unmasked region. Next, we demonstrate the efficacy of our metric through experimentation using a large dataset of synthetically generated images. Our findings indicate a strong correlation between our metric and human perception. Furthermore, we validate these results by testing \delmetric on real-world inpainting data, obtaining consistent outcomes. To the best of our knowledge, there are currently no deep feature-based metrics available for assessing inpainting quality in a reference-free manner.

\noindent
The key contributions of this work are:
\begin{enumerate}
    \item To our knowledge, this is the first reference-free metric to use deep features to compare between the inpainted regions and the other regions. Deep features align better with human perception of quality, and can provide better assessment. Furthermore, our reference free metric has the potential to resolve the impracticality of reference-based metrics due to the unavailability of the reference image in deletion.
    \item We empirically validate that \delmetric correlates well with SoTA reference-based metrics (\cref{expt-setup}), aligning well with human preference. This is also supported by a user study (\cref{user-study}) that reinforces these findings.
\end{enumerate}

\section{Related Work}

\textbf{Image Editing with Diffusion Models:} Diffusion models excel in image editing using textual prompts (and additional inputs such as masks). Notably, InstructPix2Pix~\cite{ip2p} finetunes a diffusion model using instruction prompts generated with LLMs for better textual alignment, but struggles with composability and spatial reasoning. Edit masks can alleviate this issue as shown in Refs.~\cite{glide, blended-diff, diffedit}, which utilize edit masks to perform iterative editing of objects within the image. Ref.~\cite{lomoe} improves upon this by performing zero-shot multiple object edits in a single pass. We refer the reader to Ref.~\cite{image-editing-survey} for a more comprehensive review of the domain. 

\noindent
\textbf{Object Deletion:} Viewed as inpainting, deletion aims to remove an object from the image and reconstruct the background faithfully. Generally, deletion is a challenging problem -- naive inpainting models usually fail to understand the prompt or replace the object with another object. This motivated Ref.~\cite{inst-inpaint} to solve the problem by constructing a deletion-specific dataset from the GraphVQA dataset to finetune inpainting models to deletion prompts. Ref.~\cite{magicremover} improves the performance by attaching guidance to the diffusion process.

\noindent
\textbf{Metrics for Image Quality Assessment:} Editing, inpainting, and deletion methods have been generally evaluated through image perceptual quality metrics. Most of these metrics are either fully reference-based or distribution-based, i.e., they require a reference image or a reference distribution along with the output image to measure the quality of the output. Fr\'{e}chet Inception Distance~\cite{fid}, assess the quality of generated images by comparing its distribution to ImageNet. Structural Similarity Index Measure (SSIM)~\cite{ssim} measures image degradation as a perceived change in structural information. Inspired by SSIM, ~\cite{pwiiq} propose a metric to measure the inpainting quality. SSIM-based measures may fail for images with large inpainting regions \cite{ssim-survey}. Refs.~\cite{nr-saliency-1, nr-saliency-2} introduce a reference-free saliency-based metric for inpainting. Other reference-based pixel saliency inpainting metrics have also been introduced~\cite{metric-survey}. Traditional pixel-level measures are inadequate to assess the image quality. LPIPS\cite{lpips} uses deep features of a trained network to assess image quality. CLIP Distance~\cite{inst-inpaint} evaluates object removal. Image regions from the inpainted image are extracted using a bounding box, and the CLIP similarity is measured between the inpainted region and the original textual prompt. CLIP Accuracy~\cite{inst-inpaint} utilizes the CLIP model as a zero-shot classifier for semantic label prediction -- low Top-5 accuracy indicates the object is removed.

\begin{figure*}
\centering
\hfill
\raisebox{-0.5\height}{\begin{subfigure}{0.31\textwidth}
    \captionsetup{justification=centering}
    \includegraphics[width=0.95\textwidth]{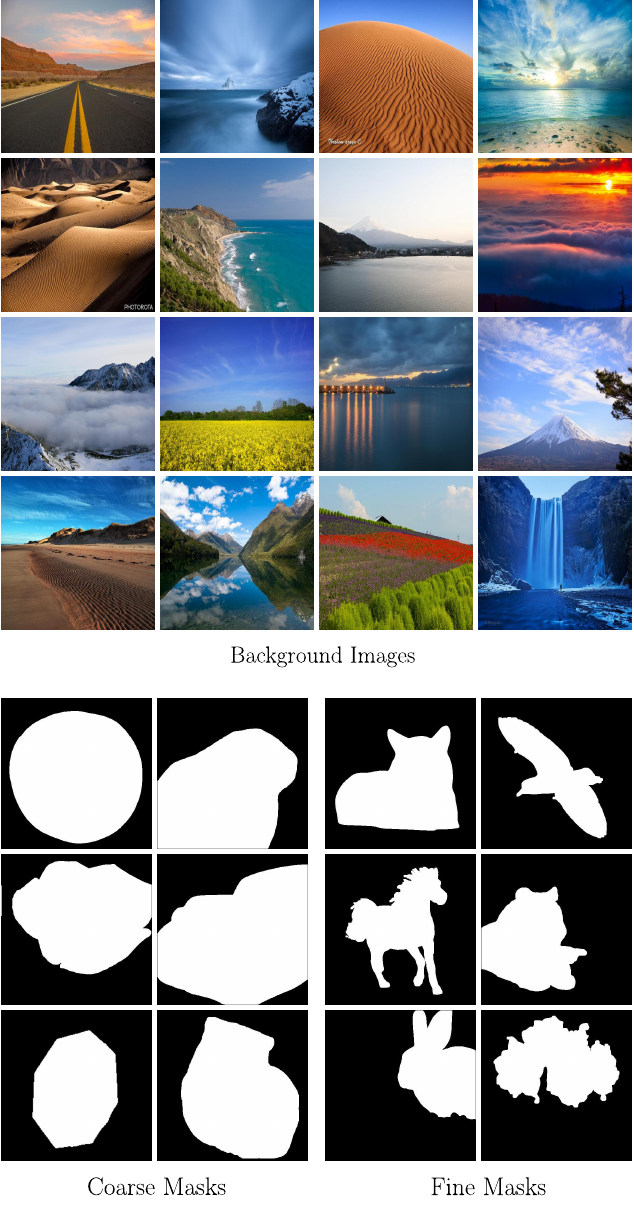}
    \caption{Sample of Background Images and Inpainting Masks in the Toy Dataset.}
    \label{fig:toy-dataset-sample}
\end{subfigure}}
\hfill
\raisebox{-0.5\height}{\begin{subfigure}{0.58\textwidth}
    \captionsetup{justification=centering}
    \includegraphics[width=0.95\textwidth]{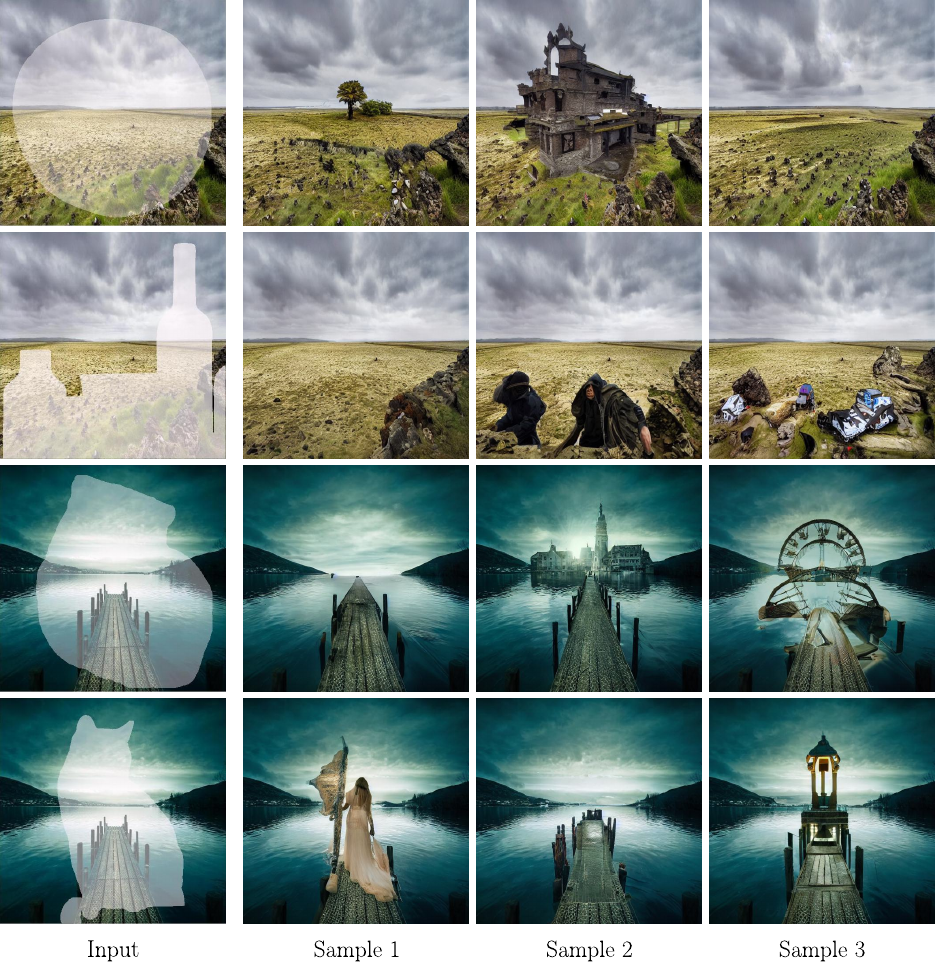}
    \caption{Samples showing the action of Stable Diffusion Inpaint upon the masked area (shown with the whitened effect) when an empty prompt (`` ") is given.}
    \label{fig:toy-dataset-application}
\end{subfigure}}
\hfill
\hfill
\caption{Toy Dataset made using background images, randomly selected masks and SD-Inpaint.}
\label{fig:toy-dataset}
\end{figure*}

\section{Proposed Metric}

\delmetric leverages feature extraction to assess inpainting quality – assessing visual saliency at a patch level rather than pixel level. Specifically, \delmetric employs a ViT~\cite{vit} trained on an image segmentation task~\cite{sam} to extract feature embeddings for the image patches. Average embeddings are calculated for the masked regions and the unmasked regions, which are then compared to score \textit{“how consistent the inpainted region is with the background regions”}. Ideally, for deletion we would want no noticeable change within the masked region compared to the background region, i.e, the similarity score is high.

Given an image $\mathbf{x}$ from image-space $\mathcal{I} \subset \mathbb{R}^{w\times h\times 3}$ ($\mathbf{x} \in \mathcal{I}$\footnote{for stable diffusion-based models, $\mathcal{I} \subset \mathbb{R}^{512\times512\times3}$}), a foreground mask $M$, a ViT feature extractor $\mathcal{E}$ and a similarity measure $\mathcal{S}$, the working of the metric can be summarized as follows:

\begin{enumerate}
    \item \textbf{Image Preprocessing}: The image $\mathbf{x}$ undergoes initial pre-processing steps including resizing and normalization according to the ViT's preprocessor~\cite{vit} to obtain $\mathbf{x}_0$. In our case, $\mathbf{x}_0 \in \mathbb{R}^{1024\times1024\times3}$.

    \item \textbf{Feature Extraction}: The ViT is used to extract features $\mathbf{z}$ from $\mathbf{x}_0$ using patches of size $p \times p$ from the image, i.e., $\mathbf{z} \in \mathbb{R}^{\frac{w}{p}\times\frac{h}{p}\times d_f}$ where $d_f$ is the embedding size for each patch. These features capture the color patterns, textures, and spatial relationships within each region~\cite{vit}. Using $\mathbf{x}_0$ and patch size $16 \times 16$, $\mathbf{z} \in \mathbb{R}^{64\times 64 \times d_f}$:
    \begin{equation}
        \mathbf{z} = \mathcal{E}(\mathbf{x}_0)     
    \end{equation}
    
    \item \textbf{Masking}: Given inpainting mask $M \in \{0,1\}^{w\times h}$, where $1$'s denote the region targeted for inpainting, we resize it to match the patch-level output, i.e, $\tilde{M} \in \{0,1\}^{\frac{w}{p}\times\frac{h}{p}}$.
    
    \item \textbf{Feature Segregation}: The features $\mathbf{z}$ are split into two disjoint sets using the mask $\tilde{M}$: \textit{masked} features $\mathbf{z}_m$ and \textit{unmasked} features $\mathbf{z}_u$.
    
    \begin{align}
        \mathbf{z}_m &= \{z_{ij} \text{ where } \tilde{M}_{ij} = 1\}_{i \in \{0, \dots, \frac{w}{p}\}, j \in \{0, \dots, \frac{h}{p}\}} \\
        \mathbf{z}_u &= \{z_{ij} \text{ where } \tilde{M}_{ij} = 0\}_{i \in \{0, \dots, \frac{w}{p}\}, j \in \{0, \dots, \frac{h}{p}\}} 
    \end{align}

    \item \textbf{Mean Features}: For both the sets, we then calculate the \textit{mean feature vector} as the average $(\,\bar{\cdot}\,)$ of features within the set, obtaining vectors $\bar{\mathbf{z}}_m$ and $\bar{\mathbf{z}}_u$. 
    
    \item \textbf{\delmetric Calculation}: The quality of inpainting is given by the similarity measure $\mathcal{S}$ between the mean masked and unmasked feature vectors.
    \begin{equation}
        \delmetric = \mathcal{S}\left(\bar{\mathbf{z}}_m, \bar{\mathbf{z}}_u\right)
    \end{equation}
\end{enumerate}

In this work, we define $\mathcal{S}$ as the cosine similarity, i.e., for \delmetric, higher values indicate better results. Additionally, we incorporate cropping during the preprocessing stage to ensure comparability between the number of patches belonging to the masked and unmasked regions. Further elaboration on this aspect can be found in \cref{real-world-dataset}.

\section{Experimental Setup}
\label{expt-setup}

In this section, we outline the experiments conducted to validate the efficacy of \delmetric as a reference-free metric for assessing inpainting quality. We first demonstrate its reliability through a simple experiment wherein we observe a consistent trend between \delmetric and perceptual similarity measured by LPIPS using the ground truth inpaintings as a baseline. Thereafter, we test the metric on a real-world inpainting scenario on the DEFACTO dataset~\cite{defacto}.

\subsection{Toy Experiment}

Traditional evaluation methods often rely on reference images, which may not be readily available. Thus, this toy experiment aims to empirically validate \delmetric for assessing image inpainting quality on a large synthetically generated dataset to check \delmetric's correlation with LPIPS.

\subsubsection{Dataset Generation}
\label{toy-expt-dataset}

We start by creating a comprehensive dataset for evaluating the effectiveness of \delmetric. Firstly, we use a collection of 4300 background images~\cite{toydata} from Flikr comprising various landscapes that can roughly be categorized as mountain, sea, desert, beach, and island. Next, we randomly select 20 masks from the PIE-Bench image editing dataset~\cite{piebench}. The masks are categorized as either \textit{coarse} or \textit{fine}, depending on whether they exhibit a general blob shape or represent a specific identifiable object, respectively. This distinction is illustrated in Figure~\ref{fig:toy-dataset-sample}.

To generate inpainted images, we employ Stable Diffusion Inpaint (SD-Inpaint)~\cite{stablediff} using randomly selected images and masks from the dataset defined above. These inpainted images utilize the input background images as their ground truth inpainting. Additionally, the inpainted images are generated with various seeds and an empty prompt (`` ") to introduce variability in the inpaintings.

The quality of inpainting varied across instances, with SD-Inpaint occasionally producing satisfactory results, as shown in Figure~\ref{fig:toy-dataset-application}, while at other times exhibiting sub-optimal performance. This variability in inpainting quality results in a diverse photo-realistic dataset comprising \textapproximate{}200,000 images with varying degrees of inpainting quality. Consequently, this dataset provides a robust foundation for evaluating our framework.

\begin{figure}[b]
    \centering
    \captionsetup{justification=centering}
    \includegraphics[width=0.9\linewidth]{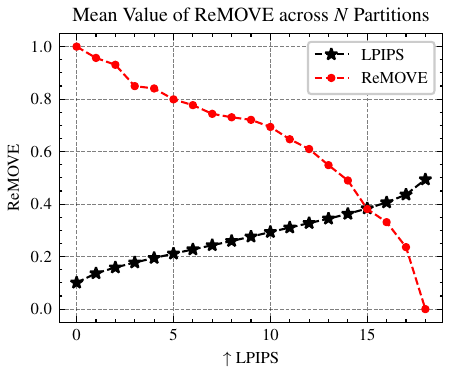}
    \caption{Result from toy experiment showing that \delmetric monotonically decreases with a decrease in inpainting quality.}
    \label{fig:toy-result}
\end{figure}

\begin{figure*}[t]
\centering
\hfill
\begin{subfigure}{0.43\textwidth}
    \includegraphics[width=0.96\textwidth]{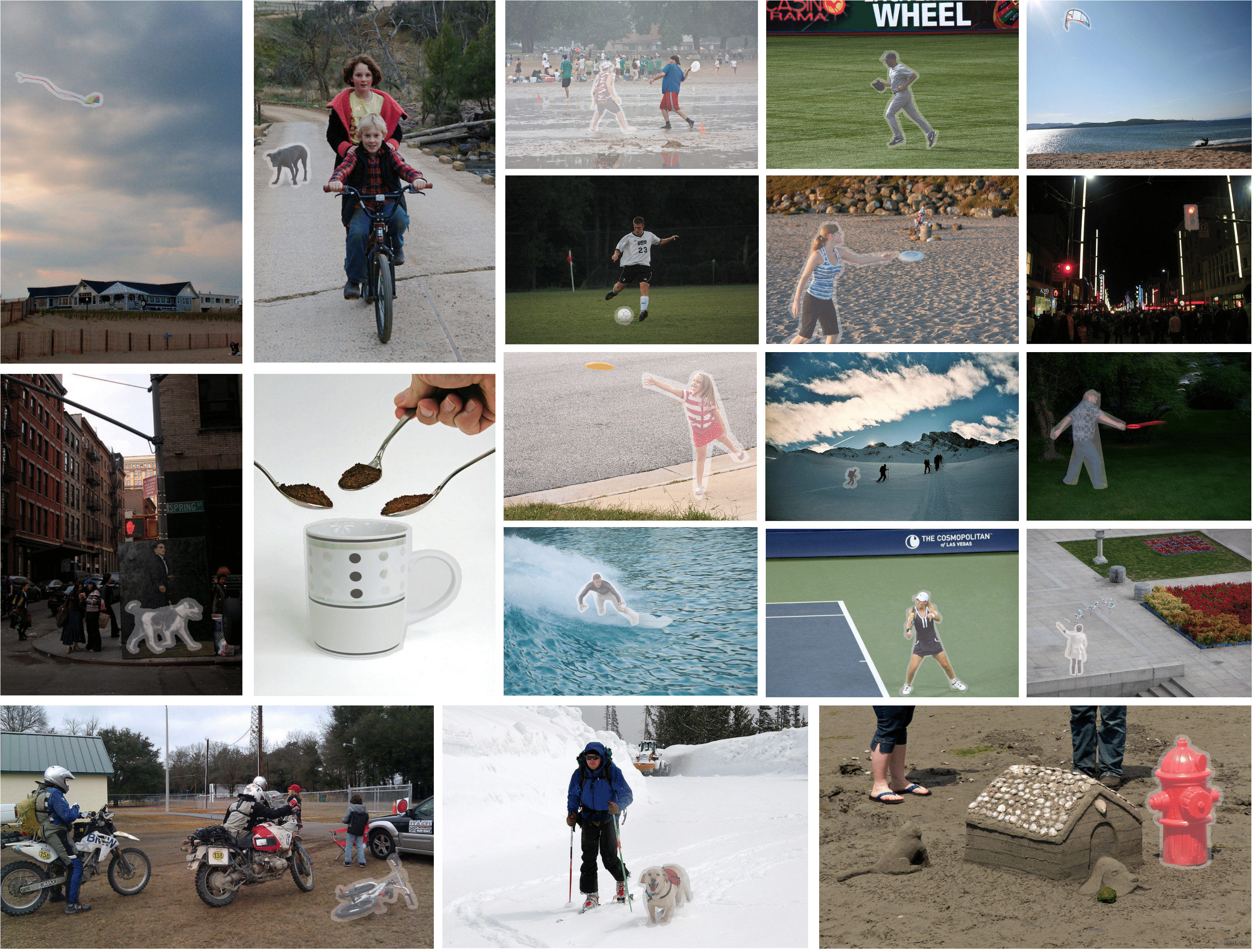}
    \caption{Real images}
    \label{fig:first}
\end{subfigure}
\hfill
\begin{subfigure}{0.43\textwidth}
    \includegraphics[width=0.96\textwidth]{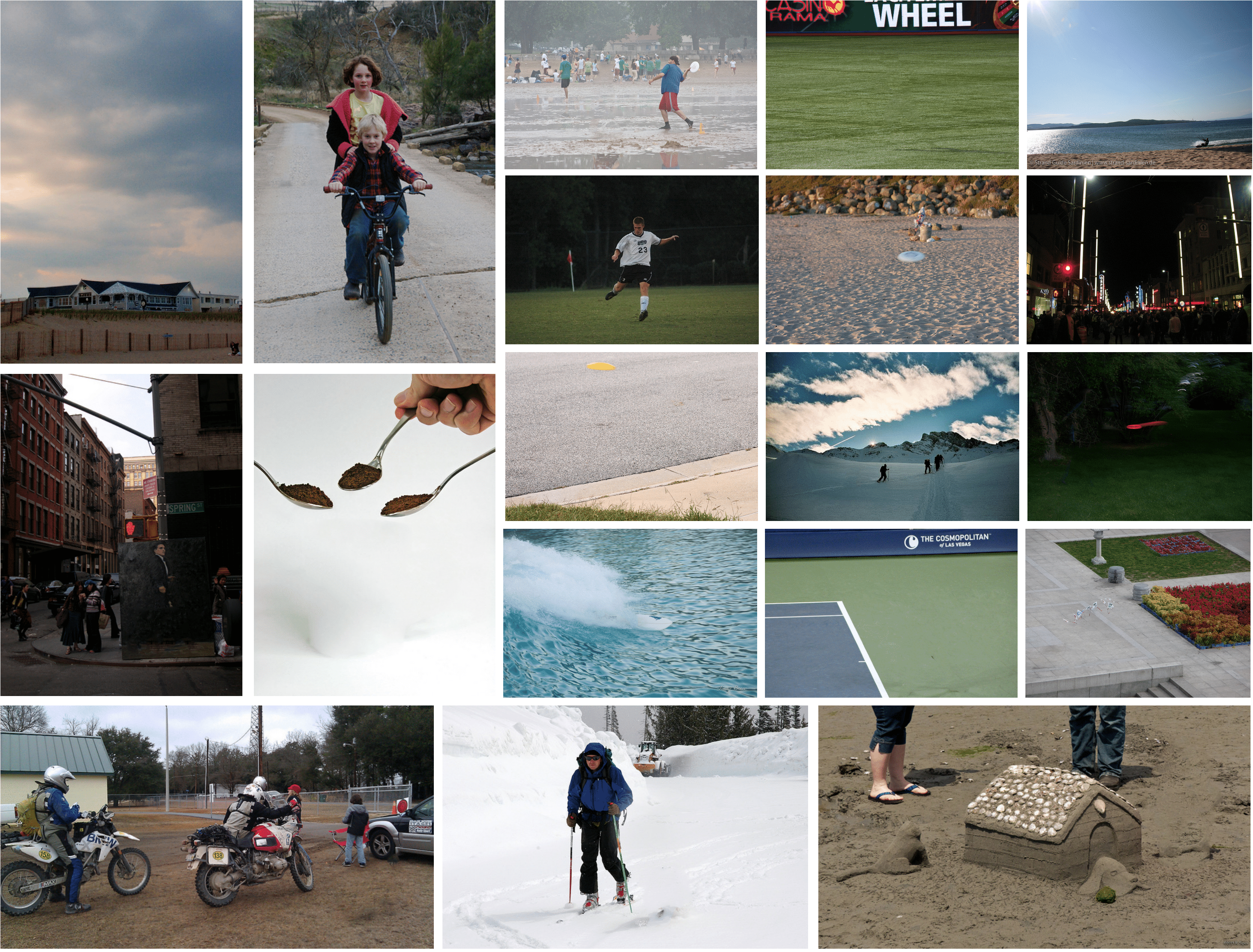}
    \caption{Ground truth inpainting corresponding to the real images}
    \label{fig:second}
\end{subfigure}
\hfill
\hfill
\newline
\newline
\begin{subfigure}{0.3\textwidth}
    \captionsetup{justification=centering}
    \includegraphics[width=0.96\textwidth]{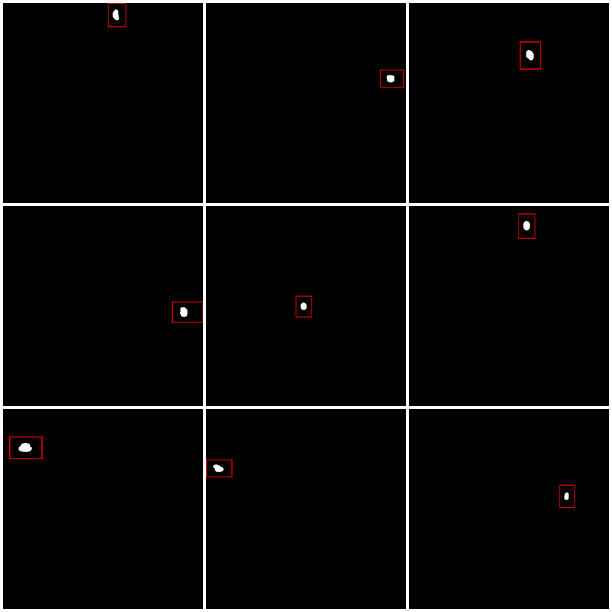}
    \caption{A random sample of \textbf{Small} masks with a box around it showing how \delmetric crops.}
    \label{fig:third}
\end{subfigure}
\hfill
\begin{subfigure}{0.3\textwidth}
    \captionsetup{justification=centering}
    \includegraphics[width=0.96\textwidth]{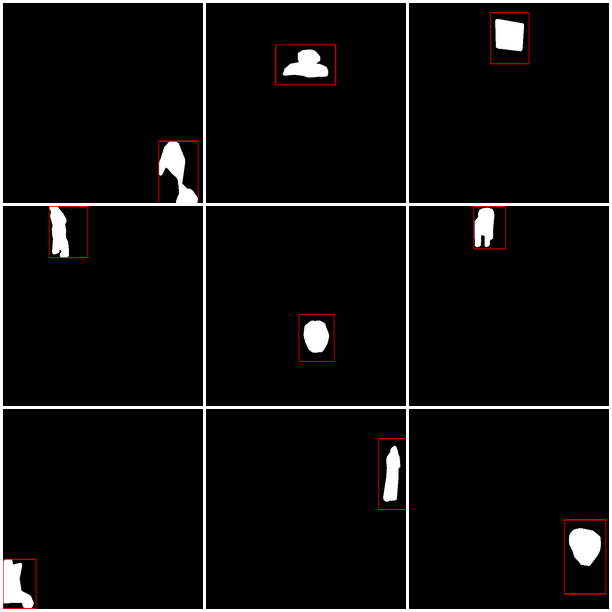}
    \caption{A random sample of \textbf{Medium} masks with a box around it showing how \delmetric crops.}
    \label{fig:third}
\end{subfigure}
\hfill
\begin{subfigure}{0.3\textwidth}
    \captionsetup{justification=centering}
    \includegraphics[width=0.96\textwidth]{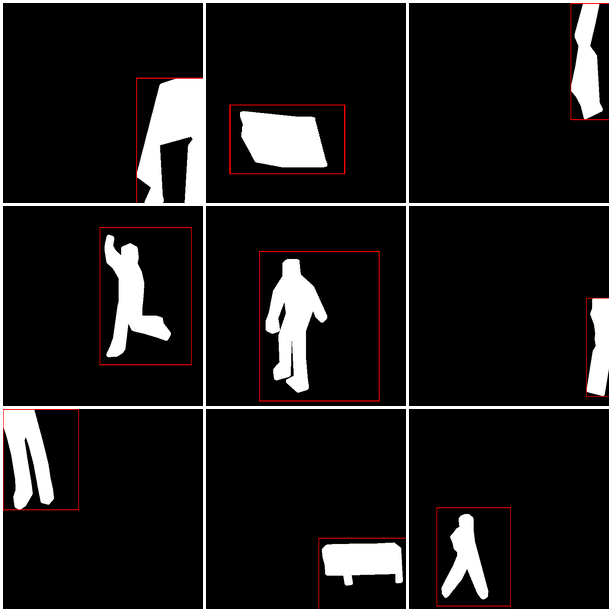}
    \caption{A random sample of \textbf{Large} masks with a box around it showing how \delmetric crops.}
    \label{fig:third}
\end{subfigure}
\caption{Samples from the object removal category of DEFACTO dataset.}
\label{fig:defacto}
\end{figure*}

\subsubsection{Empirical Validation}

To empirically validate \delmetric, we leverage LPIPS, a metric used to quantify perceptual similarity between images. A lower LPIPS score indicates a higher degree of similarity between the generated image and its ground truth, suggesting accurate inpainting. Furthermore, LPIPS has been empirically validated to closely align with human perception, making it a suitable metric for evaluating inpainting quality~\cite{lpips}.

For every generated image and its corresponding ground truth counterpart, we compute both \delmetric and the LPIPS score. Subsequently, we arrange the dataset in ascending order according to the calculated LPIPS scores, partitioning it into $N=20$ equally spaced subsets. We then determine the mean value of \delmetric within each partition. This helps facilitate a systematic analysis of inpainting quality assessment across various levels of similarity to the ground truth. Based on the findings depicted in \cref{fig:toy-result}, the decreasing trend across partitions infers a close correlation between \delmetric and the quality of inpainting, consistent with LPIPS.

In summary, this validation protocol aims to provide insights into the effectiveness of \delmetric to reliably serve as a reference-free measure for assessing inpainting quality.

\subsection{Real World Experiment}
\label{real-world-dataset}

To assess \delmetric's performance in a real-world scenario, where object masks are relatively small, we utilize the \textit{object removal} category of the DEFACTO dataset~\cite{defacto}, constructed over MSCOCO~\cite{mscoco}. Additionally, alongside LPIPS, we include comparisons with CLIPScore (CS)~\cite{clipscore}, a widely used metric for validating diffusion-based image editing performance.

\subsubsection{DEFACTO Dataset}

The DEFACTO dataset comprises \textapproximate25,000 input image, inpainting mask, and ground truth tuples. Samples from the dataset are illustrated in \cref{fig:defacto}. As CS necessitates a text prompt, we utilize BLIP~\cite{blip} to generate prompts using both the input images (resulting in CS-NR) and the ground truth inpaintings (resulting in CS-FR), where NR and FR refer to \textit{no-reference} and \textit{full-reference}, respectively~\cite{metric-survey}.

The DEFACTO dataset differs from the toy dataset (\cref{toy-expt-dataset}) primarily in two aspects: (1) it exhibits a wide variation in the types of masks present, and (2) the images within the dataset are considerably more complex compared to those in the toy experiment dataset.

\subsubsection{Empirical Validation}

\begin{figure*}[t]
    \centering
    \captionsetup{justification=centering}
    \includegraphics[width=\textwidth]{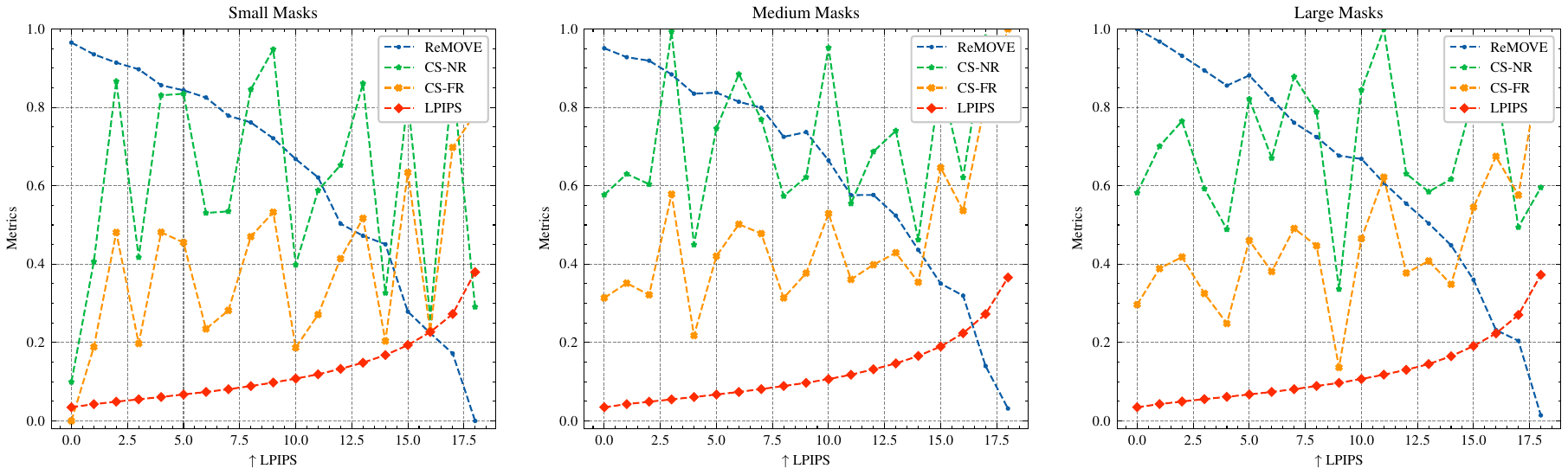}
    \caption{\textbf{Emperical results on Real Images:} \delmetric consistently aligns with LPIPS, which is the SoTA metric for assessing inpainting, while ground truth inpainted images are provided, but both CS-NR and CS-FR models are noisy and ineffective in prediction.}
    \label{fig:res-defacto-1}
\end{figure*}

Due to multiple mask sizes in the DEFACTO dataset, we partitioned it into three parts based on size: small, medium, and large, as illustrated in \cref{fig:defacto}. Following the same experimental setup as the toy experiment we sorted the images by LPIPS and computed the mean \delmetric score in each partition. This design allowed us to gain insights into the efficacy of \delmetric across various mask sizes. Furthermore, we also compare it against CS-FR and CS-NR. The results are depicted in \cref{fig:res-defacto-1} and \cref{tab:defacto-corr}. Our experiments were conducted on a GeForce RTX-3090 GPU.

\section{Discussions}

\noindent\textbf{Why does CLIPScore fail in measuring inpainting performance?}
Text captions in intricate scenes (see \cref{fig:defacto}) often lack comprehensive detail to encapsulate every object~\cite{captioning-survey}. Thus, relying solely on CLIPScore to evaluate inpainting performance may yield unreliable results. In many scenarios, ground truth data is unavailable, rendering CS-FR impractical. Consequently, CS-NR is commonly employed in practice. However, CS-NR predominantly detects the absence of the object to be inpainted without thoroughly evaluating the semantic coherence of the inpainted background. This can lead to ambiguity between object removal and replacement. The limitation stems from CS-NR's design, which prioritizes the identification of missing elements over assessing the coherence and fidelity of the inpainted region within the broader context of the entire scene. Thus, while CLIPScore offers insights, its effectiveness is constrained in complex scenes where text captions or source images may not provide sufficient information for accurate evaluation.

\noindent\textbf{Why is cropping necessary for accurately estimating inpainting performance?}
During our preliminary toy experiment, we observed promising performance from \delmetric without cropping, demonstrating a strong alignment with the expected trend with respect to LPIPS. However, as we conducted experiments with real data, we encountered a challenge posed by the varying sizes of the masks. This variability introduced complexities, especially concerning smaller masks. In such cases, we posit that evaluating the effectiveness of background inpainting may primarily focus on a localized area surrounding the mask rather than considering the entire image. This issue can be seen in Table~\ref{tab:defacto-corr} wherein the \delmetric without cropping produces implausible results for the DEFACTO dataset. Consequently, we implement a cropping strategy, as can be seen in Figure~\ref{fig:schematic}, to address this issue. We adopted a square-shaped cropping method to ensure that within the cropped regions, all masks covered a similar fraction of the area (around 30\%-50\%). This approach ensures that the number of patches belonging to the masked region is comparable to the number of patches in the unmasked region, enabling a correct evaluation of background inpainting performance across images with masks of different sizes, thus enhancing the reliability and consistency of our experimental results. 

\subsection{User Study}
\label{user-study}

This study evaluates the effectiveness of an inpainting metric in determining user preferences for inpainted images. 

\noindent
\textbf{A. Protocol:} Participants ranked a set of two inpainted images based on personal preference and rated the visual quality of inpainting. The metrics (LPIPS and ReMOVE) will independently rank the same images. Then, we measure the accuracy of the metrics in predicting the user preference.

\noindent
\textbf{B. Expected Outcome:} We hypothesize an alignment between participant and metric's preference for each inpainting pair, indicating the metric's ability to capture user preferences. 
Consistent performance across both metrics and user preferences may identify effective inpainting methods.

\noindent
\textbf{C. Results:} We conducted a user study with a set of 20 participants in the age range of 23-30. They contributed a total of 1000 data points. we observe that \delmetric agrees \textbf{74.7\%} times with the preference of the users while LPIPS agrees only \textbf{71.9\%} times. It is to be noted that LPIPS is a reference-based metric and thus has more information while evaluating the quality of inpainting than \delmetric. Thus, we conclude that in our experimental setup, \delmetric aligns more closely with the user preference than LPIPS.

\noindent
\textbf{D. Conclusion:}
This study validates the proposed inpainting metric and highlights its potential for user-centered inpainting evaluation. Insights gained can inform the development of improved evaluation methodologies.


\begin{table}
    \centering
    \captionsetup{justification=centering}
    \begin{tabular}{clccc}
        \toprule
        Dataset & Methods & \multicolumn{3}{c}{Metrics} \\
        \cmidrule(lr){3-5}
        & & $\mu$ & $\sigma$ & $\rho$ \\
        \midrule
        \multirow{2}{*}{Toy} & \delmetric (w/o crop) & 0.881 & 0.092 & -0.479 \\
         & \delmetric & 0.864 & 0.102 & -0.519 \\
        \multirow{2}{*}{Real} & \delmetric (w/o crop) & 0.528 & 0.222 & 0.145 \\
         & \delmetric & 0.851 & 0.111 & -0.515 \\
        \bottomrule
    \end{tabular}
    \caption{Ablation on the use of cropping in \delmetric include mean ($\mu$), standard deviation ($\sigma$) and correlation with LPIPS ($\rho$).}
    \label{tab:defacto-corr}
\end{table}
\section{Conclusion}
We propose \delmetric, a novel reference-free metric tailored to assess the effectiveness of object erasure post-generation in generative models like Stable Diffusion. This metric addresses the limitations of existing evaluation measures such as LPIPS and CLIPScore, especially in scenarios where reference images are unavailable, which is common in practical applications. By overcoming the challenge of distinguishing between object removal and object replacement inherent in stable diffusion models, \delmetric provides a comprehensive evaluation framework. Through empirical evaluations, we have demonstrated that \delmetric not only correlates with established metrics reflecting human perception but also captures the nuances of the inpainting process, offering a better assessment of generated outputs. We believe \delmetric will serve as a valuable tool for researchers and practitioners in evaluating and advancing image inpainting techniques, ultimately enhancing their applicability in the real-world.
{
    \small
    \bibliographystyle{ieeenat_fullname}
    \bibliography{main}
}


\end{document}